\documentclass[conference]{IEEEtran}
\IEEEoverridecommandlockouts
\usepackage{cite}
\usepackage{amsmath,amssymb,amsfonts}
\usepackage{algorithmic}
\usepackage{graphicx}
\usepackage{textcomp}

\usepackage[utf8]{inputenc}
\usepackage[font=footnotesize]{caption} 

\usepackage{float}
\usepackage{multicol}
\usepackage[numbers]{natbib}
\usepackage{multirow}
\usepackage[skip=0.33\baselineskip]{subcaption}
\def\BibTeX{{\rm B\kern-.05em{\sc i\kern-.025em b}\kern-.08em
    T\kern-.1667em\lower.7ex\hbox{E}\kern-.125emX}}
\begin{document}

\title{Out of Distribution Reasoning  by Weakly-Supervised Disentangled  Logic Variational Autoencoder\\

\thanks{This work is supported by  MoE, Singapore, Tier-2 grant number MOE2019-T2-2-040.}
}

\makeatletter
\newcommand{\linebreakand}{%
  \end{@IEEEauthorhalign}
  \hfill\mbox{}\par
  \mbox{}\hfill\begin{@IEEEauthorhalign}
}
\makeatother

\author{
  \IEEEauthorblockN{ Zahra Rahiminasab}
  \IEEEauthorblockA{\textit{School of Computer Science}\\\textit{ and Engineering} \\
    \textit{Nanyang Technological University}\\
    Singapore, Singapore \\
    rahi0004@e.ntu.edu.sg}
  \and
  \IEEEauthorblockN{Michael Yuhas}
  \IEEEauthorblockA{\textit{Energy Research Institute} \\
    \textit{Nanyang Technological University}\\
    Singapore, Singapore\\
    michaelj004@ntu.edu.sg}
  \and
  \IEEEauthorblockN{Arvind Easwaran}
  \IEEEauthorblockA{\textit{School of Computer Science} \\ \textit{ and Engineering} \\
    \textit{Nanyang Technological University}\\
    Singapore, Singapore \\
    arvinde@ntu.edu.sg}
  \linebreakand 
  
}



\maketitle

\begin{abstract}
Out-of-distribution (OOD) detection, i.e., finding test
samples derived from a different distribution than the training
set, as well as reasoning about such samples (OOD reasoning),
are necessary to ensure the safety of results generated by
machine learning models. Recently there have been promising
results for OOD detection in the latent space of variational
autoencoders (VAEs). However, without disentanglement, VAEs
cannot perform OOD reasoning. Disentanglement ensures a one-
to-many mapping between generative factors of OOD (e.g., rain
in image data) and the latent variables to which they are encoded.
Although previous literature has focused on weakly-supervised
disentanglement on simple datasets with known and independent
generative factors. In practice, achieving full disentanglement
through weak supervision is impossible for complex datasets,
such as Carla, with unknown and abstract generative factors.
As a result, we propose an OOD reasoning framework that learns
a partially disentangled VAE to reason about complex datasets.
Our framework consists of three steps: partitioning data based
on observed generative factors, training a VAE as a logic tensor
network that satisfies disentanglement rules, and run-time OOD
reasoning. We evaluate our approach on the Carla dataset and
compare the results against three state-of-the-art methods. We
found that our framework outperformed these methods in terms
of disentanglement and end-to-end OOD reasoning.

\end{abstract}


\section{Introduction}
Since machine learning models are frequently used in safety-critical applications such as autonomous driving, it is important to identify whether the results generated by machine learning models are safe. It has been shown that the distribution of training and test samples can be different~\cite{b26} and as a result, it is important to identify test samples derived from a different distribution than the training distribution, i.e., out of distribution (OOD) samples. In addition, identifying the reason behind OOD behavior (OOD reasoning) can help to provide a safe-fail mechanism to prevent or alleviate damage in safety-critical applications. 

Consider a machine learning (ML) model that controls an autonomous vehicle (AV). This model receives image data and outputs steering and throttle set points. If the model's training set was gathered in urban environments with no precipitation, both rural roads and rainy weather would be OOD. However, the risk mitigation actions for each generative factor (background, weather) should also be different, and detecting that a sample is OOD is insufficient to ensure safety. For example, samples coming from the rural road should return control to a human driver if the hazards of rural operations have not been sufficiently analyzed. Conversely, rainy samples should trigger the switch to a more conservative ML model to compensate for reduced traction if this poses less risk than returning control to an unaware human driver.

For identifying OOD samples, different models such as a variational autoencoder (VAE)~\cite{b27} can be used. A VAE learns a data distribution by encoding data in a lower-dimensional representation (latent space) and regenerating original samples from encoded representations. In general, there are two approaches for identifying OOD samples with VAEs. In the first approach, the sample likelihood is calculated in the output space of a VAE, and samples with lower likelihood are identified as  OOD samples. However, in~\cite{b12}, it is shown that OOD samples can get a higher likelihood than in-distribution samples; therefore, OOD detection in output space is unreliable. 
One solution for this problem is using the latent space of a VAE rather than output space and comparing distributions of a test sample and training samples in latent space~\cite{b6,b13,b14}. 

OOD reasoning focuses on finding the source of OOD behavior by analyzing one-to-many maps between generative factors of data and corresponding latent dimensions that encode them. Generative factors are specific data characteristics essential for data reconstruction, such as the rain intensity in an image. Disentanglement of latent space is the process of  establishing such one-to-many maps between given generative factors and their corresponding latent dimensions. However, without inductive bias in the data or model, learning disentangled latent space for a VAE is theoretically impossible~\cite{b15}. Therefore, we should train the VAE with a degree of supervision on data or apply inductive bias to the model to learn disentangled latent space.

In complex datasets such as Carla dataset~\cite{b5}, generative factors are defined at a more abstract level. In addition, not all generative factors are known, and the domain of observed generative factors can be continuous. Therefore,  providing labels based on generative factors for each image can be expensive. As a result, using match pairing~\cite{b16} for a complex dataset is more practical. However, structuring the latent space of VAE without labels and just based on partitions for more than one generative factor can be challenging. As shown in~\cite{b8}, the disentanglement performance can decrease when changes in other factors affect the learned distribution for a given factor.

Although theoretically achieving full disentanglement with weak supervision is possible, in practice, based on the size of latent space, incomplete knowledge about generative factors,  level of abstraction in defining generative factors, etc., total disentanglement may not be achieved. For example, if a fixed number of latent dimensions are selected for a rain generative factor, selected dimensions may not represent all the information regarding that factor. However, these dimensions majorly encode the rain information. Therefore, a mechanism is needed to learn partial disentanglement for complex datasets.

Logic tensor networks (LTN)~\cite{b1} distill knowledge in the network weights based on a set of rules during training. For this purpose, the loss function is defined as a set of rules. For LTN, training is the process of optimizing network parameters to minimize the satisfaction of the loss rule. Since LTN uses first-order fuzzy logic semantics (real-logic), the rules can be satisfied partially. Therefore, LTNs' characteristics make them suitable for defining partial disentanglement.

Currently, OOD detection and reasoning approaches~\cite{b6} try to achieve partial disentanglement through model-based inductive bias. However, they cannot guarantee the mapping between generative factors and specific latent dimensions.


\textbf{Our contribution:} 
To solve the aforementioned issues, we propose an  OOD reasoning framework that consists of three phases: data partitioning, training OOD reasoners, and run-time OOD reasoning. Data partitions are formed based on observed values for generative factors, and OOD reasoners (latent dimensions of a weakly-disentangled VAE) are designed with match pairing supervision and LTN.
 Using the LTN version of a VAE allows us to define disentanglement formally based on given data partition samples. Inspired by~\cite{b16} we define the \emph{adaptation} and \emph{isolation} rules for achieving disentanglement. The \emph{adaptation} rule ensures that the change in generative factor values is reflected in the distribution learned for the corresponding latent dimensions. The \emph{isolation} rule guarantees that the change in a given factor is only reflected in its corresponding latent dimensions. Since LTN uses first-order fuzzy logic semantics (real-logic), \emph{adaptation} and \emph{isolation} rules can be partially satisfied. As a result, the VAE can achieve a proper level of disentanglement even when latent space size is small, and some generative factors are not observed during training. Finally, we use the corresponding dimension for a given factor to identify OOD samples based on a given factor. We show the effect of defined constraints on disentanglement by visualization, and also mutual information~\cite{b28}. We also show that our approach 
achieves an AUROC of 0.98 and 0.89 on the Carla dataset for rain and city reasoners, respectively.

\section{Background}\label{sc:Back}
Our framework is built based on three concepts. In this section, we introduce these concepts. 

\textbf{Variational autoencoder:} 
 A VAE is a machine learning model formed by two attached neural networks: the encoder and the decoder. Given an input $x$,  encoder $q_{\phi}(z|x)$ with parameters $\phi$ maps the input to latent representation $z$. The decoder $p_{\theta}(x|z)$ with parameters $\theta$ regenerates data from  $z$ representation.  Equation \ref{eq:ELBO} shows the ELBO loss of a VAE.

\begin{equation}
    loss=E_{q_{\phi}(z|x)} [\log \; p_{_{\theta}}(x|z)] -KL(q_{\phi}(z|x) || p(z))
    \label{eq:ELBO}
\end{equation}

The first and second terms are reconstruction loss and regularization losses, respectively. The reconstruction loss ensures that the distribution learned for data reflects the main factors required for data reconstruction. The regularization loss evaluates the similarity between the learned distribution and the prior distribution by using KL-divergence between learned and prior (usually standard Gaussian distribution $\mathcal{N}(0,1)$) distributions~\cite{b23}.

\textbf{Logic tensor network:}
A logic tensor network (LTN) uses logical constraints to distill knowledge about data and a model in model weights during training. The knowledge is formally described by first-order fuzzy logic named real logic. Real logic uses a set of functions, predicates, variables, and constants to form logical terms. These elements, alongside operators such as negation, conjunction, etc.,  form the syntax of real logic. 
The fuzzy semantic is defined for real logic so that the rules can be satisfied partially. The operators are semantically defined based on product real logic. Table \ref{tab:Reallog}, summarizes the operator definition. In this table $a$, $b$, and $a_1,...,a_n $ are predicates with values  that fall in $[0,1]$.  Learning for a logic tensor network is the process of finding a set of parameters that maximize the satisfaction of rules or minimizing the satisfaction of a loss rule defined by real logic~\cite{b1}. 
\begin{table}[]
\centering
\begin{tabular}{|l|c|c|}
\hline
\multicolumn{1}{|c|}{\textbf{Symbol}} & \textbf{Definition}                                & \textbf{Explanation}                                                               \\ \hline
$\sim: N(a)$                          & 1-a                                                & \begin{tabular}[c]{@{}c@{}}Standard \\ negation\end{tabular}                       \\ \hline
$\wedge: T(a,b)$                      & $a.b=\sum{x_iy_i}$                                 & \begin{tabular}[c]{@{}c@{}}Product t-Norm. \\ (.) is inner\\  product\end{tabular} \\ \hline
$\vee:S(a,b)$                         & a+b-a.b                                            & Dual t-conorm                                                                      \\ \hline
$\rightarrow: I(a,b)$                 & 1-a+a.b                                            & \begin{tabular}[c]{@{}c@{}}Reichenbach\\  implication\end{tabular}                 \\ \hline
$\exists:A_{pM}(a1,..,a_{n})$         & $\sqrt{(\frac{1}{n} \sum_{i=1}^{|n|}a_i^2)}$       & \begin{tabular}[c]{@{}c@{}}Generalized \\ mean\end{tabular}                        \\ \hline
$\forall$ : $A_{pME}(a_1,..,a_{n})$   & $1-\sqrt{(\frac{1}{n} \sum_{i=1}^{|n|}(1-a_i)^2)}$ & \begin{tabular}[c]{@{}c@{}}Generalized  \\ mean w.r.t.\\  the error\end{tabular}   \\ \hline
\end{tabular}
\caption{Real logic fuzzy operations and their definition}
\label{tab:Reallog}
\end{table}

\textbf{Weakly supervised disentanglement:}
Disentanglement is defined by two concepts: consistency and restrictiveness. Given a generative factor $s$ encoded in dimensions $i \in I$ of latent space, consistency means changes in the distributions of dimensions outside the specified set ($\overline{I}$) do not affect given factor $s$. Restrictiveness means other factors $s' \in S \setminus \{s\}$ are immune to changes in the distributions of specified dimensions ($i \in I$) that encode generative factor $s$~\cite{b16}. We can use different levels of weak supervision to attain disentanglement, such as restricted labeling, rank pairing, and match pairing~\cite{b16}. Restricted labeling has the highest degree of supervision among these three approaches as it requires information, usually in the form of labels about generative factors of a sample in combination with samples for training~\cite{b17,b18,b21}. The rank pairing has a lower degree of supervision as it requires an auxiliary variable showing the relation between samples' generative factors alongside the training samples~\cite{b7}. Finally, match pairing has the weakest degree of supervision among the three approaches as it only receives partitions of data during training that share either the same values of a specific generative factor or only differ in specific numbers of generative factors~\cite{b8}.

\section{Related work}\label{sc:Related}

Approaches that use VAE as an OOD detector and reasoner analyze either the output space or latent space of a VAE to identify OOD samples. As shown in~\cite{b12}, OOD detection in output space can lead to wrong results as OOD samples can have higher likelihoods than in-distribution samples. As a result,  recent approaches focus on OOD detection, and reasoning in the latent space~\cite{b6,b13,b14}. While OOD detection in latent space leads to promising results, for OOD reasoning, there should be a mechanism to correspond generative factors to the latent dimensions that represent them. Disentangling the latent space of a VAE is such a mechanism.

Based on the degree of supervision, the approaches that provide disentanglement for VAEs are categorized as supervised, weak/semi-supervised, and unsupervised. As mentioned in~\cite{b15}, achieving disentanglement without any form of supervision and enforcing inductive bias on model and data is theoretically impossible; as a result, in the past years, the focus has shifted to learning disentanglement method based on weakly-supervised or supervised approaches. The weak supervision method is categorized into three categories, restricted labeling, rank pairing, and match pairing~\cite{b16}.

\textbf{Restricted labeling:} 
Approaches that use restricted labeling get data alongside labels based on generative factors as input. The loss function of VAE is modified to learn disentangled latent space based on provided labeled data~\cite{b17,b18}. The model obtained by this kind of weak supervision is the conditional variational autoencoder (CVAE)~\cite{b19}. At inference time, based on provided labels, samples that do not belong to training distributions are identified as OOD. An important issue of the CVAE approach is the need for data labels for all input samples. Providing data labels for complex datasets can be expensive. In addition, the basic assumption of CVAE approaches is that the factors of variation are independent; therefore, there is no correlation between labels for generative factors. However, 
When the dataset is complex and generative factors are specified more abstractly, such correlation exists~\cite{b20}. In the same line of research in~\cite{b21}, information that can be specified in different forms, for example, labels, is used to achieve disentanglement. Two VAEs are attached in a uniform architecture to provide disentanglement. One variational autoencoder is used to recreate the information from latent space and another to recreate input from latent space. Learning two VAEs can be resource-intensive.  

\textbf{Rank pairing:} Rank pairing approaches specify the relation between generative factors of pair of samples based on an auxiliary variable. In~\cite{b7} a similarity score, which can be a binary or real value,  is used to describe the relationship between generative factors of pair of samples. The similarity score is augmented in the loss function to ensure a fixed dimension in latent space only encodes information based on similarity scores of a corresponding generative factor. While defining a similarity score for one generative factor with discrete values is possible, for several generative factors that can change simultaneously, defining a similarity score that can effectively distinguish between different factors and disentangle them can be challenging. 

\textbf{Match pairing:} 
Match pairing approaches only use pairs or groups of samples for supervision during training. This approach is more applicable to complex datasets than restricted labeling and rank pairing as it requires less supervision compared to them. In~\cite{b8}, a pair of samples that only differ in a limited number of unknown generative factors are fed to a VAE.
Then, the latent dimensions are divided into higher and lower bins based on KL-divergence between learned distributions in latent space for the given pair of samples. Dimensions in lower bins are averaged by the methods in~\cite{b10} or~\cite{b11}. The distributions in higher bins remain intact, and the dimensions that fall into higher bins encode changing generative factors. While this approach provides promising results, the factors and corresponding latent dimensions are not identified before training. As a result, this approach is unsuitable for OOD reasoning in the presence of more than one generative factor. Besides, their experiment indicates the poor performance of the approach when several generative factors change simultaneously.

\textbf{Model-based disentanglement:}
In~\cite{b6}, model-based inductive bias is used to disentangle latent space. The $\beta$ and size of latent space are optimized based on disentanglement metric MIG~\cite{b9} using  Bayesian optimization (BO)~\cite{b29}.
 After training a VAE with the selected hyper parameters, scenes of time-stamped data that differ only on specific generative factors are fed to the model. Dimensions with the highest average KL-divergence difference for subsequent pairs of samples in the entire scene are selected as OOD reasoners. Since the correspondence of a factor to a latent dimension is not established during training, the dimension that encodes a specific factor can be different at test time.

\textbf{Constraint-based disentanglement:} Approaches, such as~\cite{b24} use semantic information of data to disentangle a  latent space. This approach uses the relation between each pair of numbers (such as equality of numbers) in MNIST~\cite{b25}  for achieving disentanglement. However, this approach focuses on the semantic characteristic of input data rather than relations between data distributions of generative factors in latent space to create a disentangled latent space. As a result, these approaches can not be generalized for different datasets.

\section{Method}

Figure \ref{fig:overview} shows the proposed framework. Our framework consists of three phases: data partitioning, training OOD reasoners, and OOD reasoning in latent space. In the data partitioning step, we divide the samples based on observed generative factors into partitions. We also fix the latent dimensions that encode the information of a specific generative factor. Based on data partitions and selected latent dimensions, in the next step,  we define rules that are necessary for the training of the VAE (\emph{Recloss, Regloss}) and rules that ensure disentanglement of a given factor in specified dimensions in latent space (\emph{Adaptloss, Isoloss}). We define the loss rule as an augmentation of these four types of rules. We then train the LTN version of the VAE to minimize satisfaction of the defined loss function rule to achieve disentangled latent space. Finally, in the run-time phase, we use the disentangled latent space to identify samples that are OOD with respect to a specific factor. In the following subsections, we explain each phase in detail.

\begin{figure}
    \centering
    \includegraphics[width=0.5\textwidth]{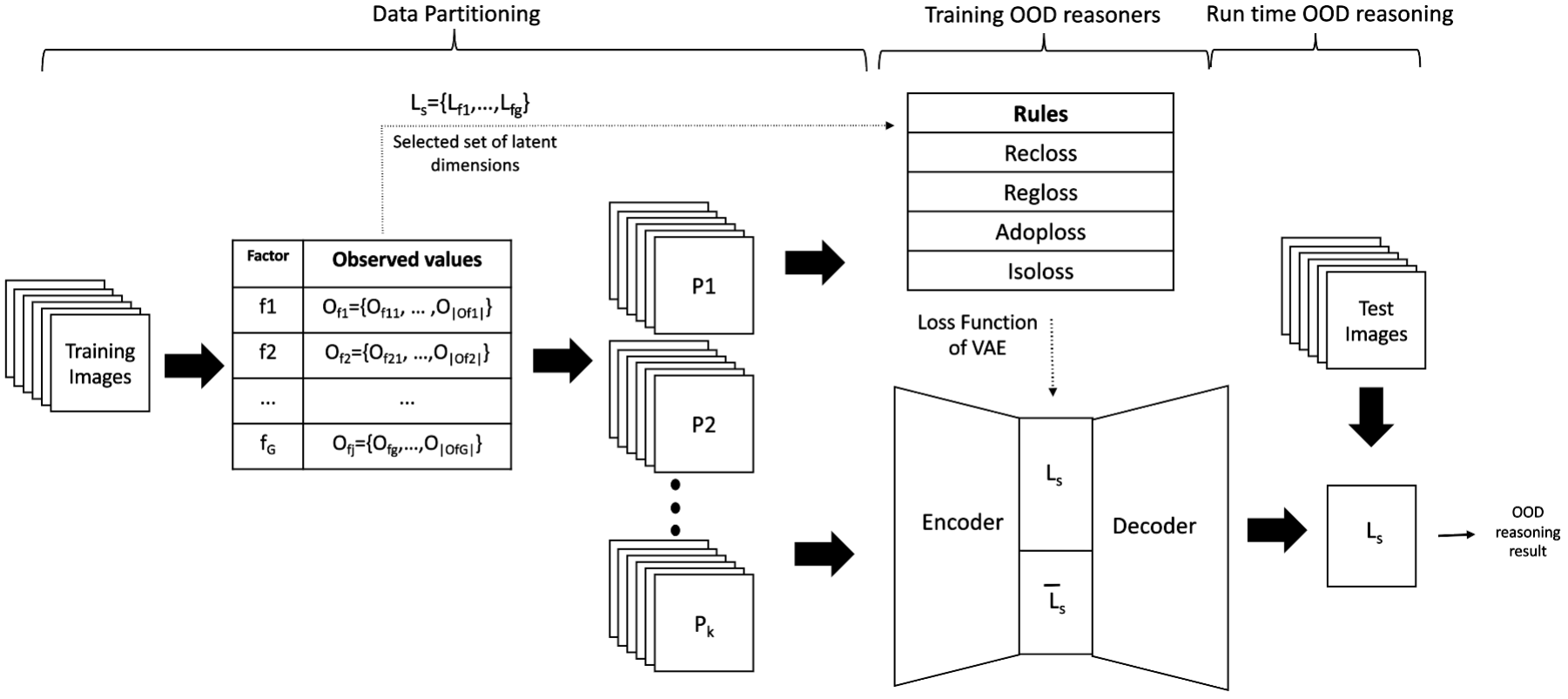}
    \caption{Overview of proposed framework}
    \label{fig:overview}
\end{figure}

\subsection{Data partitioning}

Consider an image dataset $X$ and set of observed generative factors $F=\{f_1,..., ... ,f_G\}$ where $G$ is the number of observed generative factors.  Each factor can take values from set $ O_{f_{i}}=\{o_{i1},... ,o_{ij}, ... ,o_{|O_{f_{i}}|} \}$ where $o_{ij}$ is the $j$  observed value for generative factor $f_{i}$ during training and   $|O_{f_{i}}|$ is the number of all observed values for generative factor $f_{i}$ during training. The observed value for generative factors can be discrete or continuous. When factors are continuous, values are discretized into the  bins. In this case each $o_{ij}$ shows the assigned values  to the bins that are discretizing the  factor $f_i$. 

Based on the combination of observed values for observed generative factors we then form data partitions $P_{k}=\{x \in X|(x_{f_1},...,x_{f_G})=(o_{1j},...,o_{Gj'}) \}$ where $x_{f_i}$ shows the value of generative factor $f_i$ for sample $x$ and $k \in K =|O_{f_1}| \times ... \times |O_{f_G}|$.  The number of partitions (K) equals the combination of values observed for the given factors.

As an example consider two generative factors $f_{rain}$ and $f_{city}$ for rain  and city, respectively. City  shows the number assigned to different cities where the image is captured.  Assume, for rain, low and moderate rain intensities ($O_{f_{rain}}=\{LR,MR\}$) and for city, city three and city four ($O_{f_{city}}=\{SC3,SC4\}$) are observed. Then  four partitions are formed based on the combination of values for rain and city generative factors. One example of such partitions is  $P_1=\{x \in X | (x_{f_{rain}},x_{f_{city}})=(LR,SC3)\}$.

For each generative factor $f_i$, we select one or more dimensions in latent space to encode it. We define $L_{f_i} \subset \mathcal{L}$ as a set containing all dimensions that encode generative factor $f_i$ where $\mathcal{L}$ is the set of all dimensions of latent space. 
Consider a set $V_{f_i}= \{(P_k, P_{k'}) \in P \times P\}$
where $P_{k}=\{x \in X|(x_{f_1},...,x_{f_G})=(o_{1j}, ... ,o_{ij}, ... ,o_{Gj}) \}$ and  $ P_{k'}=\{x \in X| (x_{f_1},...,x_{f_G})=(o_{1j},... ,o_{ij'}, ... ,o_{Gj})$ and  $o_{ij} \neq o_{ij'}$. This set contains pairs of partitions that only differ in one generative factor.
For example, partitions $P_1=\{x \in X|(s_{f_{rain}},s_{f_{city}})=(LR,SC3) \}$ and  $P_2=\{x \in X|(s_{f_{rain}},s_{f_{city}})=(MR,SC3) \}$ belong to $V_{f_{rain}}$.

\subsection{Training OOD reasoners}

LTN~\cite{b1} based models distill knowledge during training in network weights. Since the knowledge is described as logical rules, they also provide interpretability during training. In addition, since we use real logic, which is first-order fuzzy logic, we can also define partial satisfaction for different rules. As a result, we train a weakly-disentangled logic tensor VAE (WDLVAE) to achieve partially a disentangled latent space.

Since training in LTN  translates to optimizing the network parameter that maximizes the satisfaction of negation of the loss rule, in the first step, we define the loss rule based on a set of real logic rules. Given  image tuples $(x_1,...,x_K)$ Where $x_1 \in P_1,... ,x_K \in P_K$. Each image has width $W$, height $H$ and number of channels $C$ ($x_i \in R^{H \times W \times C}$).  Consider the hyper parameter $n=|\mathcal{L}|$ the size of latent space. Then the functions and predicates that are used for defining rules are defined as follows for each image $x_k$ in image tuples.

\textbf{Functions:}
Functions are defined in first half of table \ref{tab:FuncsPreds}. The \emph{Enc} function  is realized as an encoder network with latent space of size $n$ and parameter $\phi$ . This function takes the batch of images and maps them to the latent space of a VAE by learning the mean ($\mu_k$) and the logarithm of variance ($log(\sigma^2)_k=lg_k$) of the distribution of data sample $x_k$. The \emph{Samp} function takes the tuple ($\mu_k$, $lg_k$) from the \emph{Enc} network and returns a sample $z_k$ by applying the reparameterization trick~\cite{b27}. The \emph{Dec} function reconstructs the image based on the sample generated by \emph{Samp}. The \emph{Dec} function is realized as a decoder network with parameter $\theta$. As the \emph{Enc} function returns tuples of ($\mu_k$, $lg_k$), the functions \emph{Mean}, and \emph{Logvar} are used for extracting $\mu_k$ and $lg_k$, respectively. 

\begin{table}[]
\begin{center}
\begin{tabular}{|c|c|c|}
\hline
\textbf{Function} & \textbf{Domain and Range}                                                   &\textbf{ Definition}                                                                                                                                       \\ \hline
Enc    & \begin{tabular}[c]{@{}c@{}}$R^{H  \times W \times C} \longrightarrow$  \\ $R^{n} \times R^{n}$\end{tabular} & \begin{tabular}[c]{@{}c@{}} $Enc_{\phi}(x_k)=(\mu_k,lg_k)$\end{tabular}                                            \\ \hline
Dec    & \begin{tabular}[c]{@{}c@{}}$R^{ n} \longrightarrow$\\$ R^{ H \times W \times C}$\end{tabular}  & \begin{tabular}[c]{@{}c@{}} $Dec_{\theta}(z_k)=\hat{x_k}$  \end{tabular} \\ \hline

Samp                                  & \begin{tabular}[c]{@{}c@{}}$R^{ n } \times R^ {n} \longrightarrow$\\ $R^{ n }$ \\ \end{tabular} & \begin{tabular}[c]{@{}c@{}}  $Samp((\mu_k,lg_k))=z_k$\end{tabular}        \\ \hline

Mean                        & \begin{tabular}[c]{@{}c@{}}$R^{ n } \times R^ {n} \longrightarrow$\\$ R^{ n } $\end{tabular}    & $Mean((\mu_k,lg_k))=\mu_k$                                                                              \\ \hline
Logvar                      & \begin{tabular}[c]{@{}c@{}}$R^{n } \times R^ {n} \longrightarrow$\\ $R^{ n } $\end{tabular}    &    $ Logvar((\mu_k,lg_k))=lg_k$                                                                              \\ \hline

\textbf{Predicate} & \textbf{Domain and range}  & \textbf{Definition}      \\ \hline

Rec   & \begin{tabular}[c]{@{}c@{}}$R^{H \times W \times C}$ \\ $\times R^{H \times W \times C}$ \\ $\longrightarrow $  $[0,1]$

\end{tabular} & \begin{tabular}[c]{@{}c@{}} $Rec(x_k,\hat{x_k})=$ \\ 
   $\frac{1}{H*W}[(x_k - \hat{x_k}) ^2]$\end{tabular}                                         \\ \hline
KLU & \begin{tabular}[c]{@{}c@{}}
$R^{n} \times R^{n}$ \\ $\longrightarrow[0,1]$\end{tabular}                              & \begin{tabular}[c]{@{}c@{}}  $KLU(\mu_k,lg_k)=$\\$    [\frac{-lg_k}{2}+ \frac{e^{lg_k}+
     \mu_k^ 2)}{2}$  $-\frac{1}{2}]$\end{tabular} \\ \hline
KLT & \begin{tabular}[c]{@{}c@{}}
$R^n \times R^n \times$ \\ $R^n  \times R^n$ \\ $\longrightarrow$ $[0,1]$\end{tabular}                   & \begin{tabular}[c]{@{}c@{}} $KLT(\mu_k,lg_k,\mu_k',lg_{k'})=$\\$    [\frac{(lg_{k'}-lg_k)}{2}+ \frac{e^{lg_k}+
     (\mu_k-\mu_{k'})^ 2}{2*e^{lg_{k'}}}$ \\ $-\frac{1}{2}]$\end{tabular}                                  \\ \hline

\end{tabular}
\caption{Functions and predicates that are used for defining rules.}
\label{tab:FuncsPreds}
\end{center}
\end{table}
\textbf{Predicates:} Predicates are defined in the second half of table \ref{tab:FuncsPreds}. There are two differences between functions and predicates. First, the predicate maps any input to $[0,1]$. Second, the logical operator only can be defined over predicates. As shown in the table,  the \emph{Rec} predicate indicates the reconstruction loss between the original and reconstructed image. \emph{KLU} predicate shows  the KL-divergence between a prior distribution  (standard Gaussian distribution($\mathcal{N}(0,1)$)) and the given distribution. \emph{KLT} indicates the KL-divergence between two distributions. 
Since the output of \emph{Rec}, \emph{KLU}, and \emph{KLT} can be outside $[0,1]$, we will normalize them for each batch of images that pass through a  network for training.

\textbf{Rules:} Rules are used to define a loss function to train a VAE with disentangled latent space. The operators that are used in  rules are defined in table \ref{tab:Reallog} in section  \ref{sc:Back}. We define four types of rules: \emph{Recloss} , \emph{Regloss}, \emph{Adaptloss} and \emph{Isoloss}.

The first two rules are necessary to train a VAE. We referred to~\cite{b33} to define these rules.  Equation \ref{eq:ReclossP} shows the $Recloss_{P_k}$ rule.

\begin{equation}
    \begin{gathered}
    Recloss_{P_k}: \forall{x_k \in P_{k}} \\ (Recloss(x_k,Dec_{\theta}(Samp(Enc_{\phi}(x_k)))))
\end{gathered}
\label{eq:ReclossP}
\end{equation}

This rule penalizes the Euclidean squared distance between the original sample and reconstructed sample for samples that belong to partition $P_k$. The total \emph{Recloss} over all partitions is defined by the general mean ($\exists$) operator. Equation \ref{eq:Recloss} shows how the rules for each partition are combined.

\begin{equation}
    \begin{gathered}
    Recloss: \exists k \in K \; (Recloss_{P_k})
\end{gathered}
\label{eq:Recloss}
\end{equation}

This rule aggregates the reconstruction loss obtained for each partition $P_k$ based on the general mean. 

Equation \ref{eq:ReglossP} shows the $Regloss_{P_k}$ rule. 
 
\begin{equation}
\begin{gathered}
   Regloss_{P_k}: \forall x \in P_{k} \\ 
   (KLU(Mean(Enc_{\phi}(x)),  Logvar(Enc_{\phi}(x))))   
\end{gathered}
\label{eq:ReglossP}
\end{equation}

This rule penalizes the KL-divergence between distribution learned for a sample that belongs to partition $P_k$ and prior distribution (standard Gaussian distribution $\mathcal{N}(0,1)$).

Similarly, the rule defined in equation \ref{eq:Regloss} aggregates the regularization loss over partitions.

\begin{equation}
    \begin{aligned}
    Regloss: \exists k \in K \; (Regloss_{P_k})
\end{aligned}
\label{eq:Regloss}
\end{equation}


We define the last two rules to ensure the dimensions $L_{f_i}$ of latent space encode only information of factor $f_i$ by using partitions that belong to $V_{f_i}$. 

The third type of rule ensures adaptability of a distribution to different values of generative factor $f_i$ encoded in dimensions $L_{f_i}$. In other words, we ensure the distribution learned in dimension $L_{f_i}$ is sensitive to changes in values of generative factor $f_i$. 
We call this loss \emph{Adaptloss}. For a given sample's tuple $(x,y)$ from pair of partitions $(P_k,P_{k'})$ that only differ in one generative factor equation \ref{eq:Adoptf} shows the \emph{Adaptloss} rule. 

\begin{equation}
\begin{gathered}
       Adaptloss_{(P_k,P_k')}: \sim (\forall (x, y) \in (P_k, P_{k'}) \\
   (KLT(Mean_{L_{f_i}}(Enc_{\phi}(x)),Logvar_{L_{f_i}}(Enc_{\phi}(x)),\\
    Mean_{L_{f_i}}(Enc_{\phi}(y)),Logvar_{L_{f_i}}(Enc_{\phi}(y)))))
\label{eq:Adoptf}
\end{gathered}
\end{equation}
Where $Mean_{L_{f_i}}$ and $Logvar_{L_{f_i}}$ are dimensions $L_{f_i}$ of mean and logarithm of variance. The KL-divergence will be averaged over all dimensions in $L_{f_i}$. This equation penalizes learning the same distribution for different values of generative factor $f_i$. If we assume the prior distribution is $\mathcal{N}(0,1)$, this rule leads to learning the Gaussian mixture model~\cite{b2} in dimensions corresponding to the given generative factor. As a result, it also alleviates the problem of topology mismatch~\cite{b3}. Topology mismatch occurs when the prior distribution can not represent the distribution of data properly due to the complexity of data.

In equation \ref{eq:Adopt} we aggregate the rules over pairs of partitions that only differ on a given generative factor $f_i$. 

\begin{equation}
\begin{gathered}
 Adaptloss_{f_i}: \bigwedge_{(P_k, P_k') \in V_{f_i}} Adaptloss(P_k,P_k')
 \end{gathered}
 \label{eq:Adopt}
\end{equation}

Finally \emph{Isoloss} ensures the change in generative factor $f_i$ is only encoded in dimensions that belong to $L_{f_{i}}$ and does not affect other dimensions of latent space ($\overline{L_{f_i}}=\mathcal{L} \setminus L_{f_i}$). In other words, other dimensions ($\overline{L_{f_i}}$) are insensitive to change in generative factor $L_{f_i}$. Equation \ref{eq:Isop} shows this rule for a given pair of partitions $(P_k,P_{k'})$ that only differ in one generative factor.

\begin{equation}
  \begin{gathered}
     Isoloss_{(P_k,P_k')}:  \forall (x,y) \in (P_k, P_{k'}) \\
(KLT(Mean_{\overline{L_{f_i}}}(Enc_{\phi}(x)),Logvar_{\overline{L_{f_i}}}(Enc_{\phi}(x)),\\
Mean_{\overline{L_{f_i}}}(Enc_{\phi}(y)),Logvar_{\overline{L_{f_i}}}(Enc_{\phi}(y))))
\end{gathered} 
\label{eq:Isop}
\end{equation}

This equation, penalize the change in distribution of dimensions $\overline{L_{f_i}}$. 
Similar to $Adaptloss$ rules, the $isoloss$ rules are aggregated over pair of partitions that only differ on given generative factor $f_i$, as shown in equation \ref{eq:Iso}.

\begin{equation}
\begin{gathered}
  Isoloss_{f_i}: \bigwedge_{(P_k, P_k') \in V_{f_i}}  Isoloss(P_k,P_k')
 \end{gathered}
 \label{eq:Iso}
\end{equation}

We use fuzzy semantics of operators as mentioned in table~\ref{tab:Reallog} to translate satisfiability of rules that are mentioned in equations \ref{eq:ReclossP} -  \ref{eq:Iso} to real numbers. 
The total loss function is defined as an average of rules over the number of rules by equation \ref{eq:loss}. 

\begin{equation}
\begin{gathered}
      Loss({P_1},...,P_k)= \frac{1}{(2+2|F|)} \times \\
       [Recloss+Regloss+\sum_{f_i\in F}(Adaptloss_{f_i}+Isoloss_{f_i})]
\end{gathered}
 \label{eq:loss}   
\end{equation}

The training is then done to find $\phi$ and $\theta$ parameters that minimize satisfaction of the Loss rule.

\subsection{OOD reasoning}

We use dimensions $L_{f_i}$ as OOD reasoners for generative factor $f_i$. We then form data clusters in dimension $L_{f_{i}}$ by using the k-means~\cite{b4} algorithm. Finally, we approximate the Gaussian mixture model~\cite{b2} by using cluster centers. If the probability of membership of test sample $x_t$ is less than the specific threshold $\tau_{f_i}$, we identify the test sample as an OOD sample with respect to the generative factor $f_i$.

\section{Experiments and Results}

We applied our approach on the complex Carla dataset~\cite{b5}. We used a desktop computer with an Intel i9 processor,  Geforce RTX $3080$ GPU, and   $64$ GB memory for experiments. 

The training of the model is performed offline, outside the cyber-physical system (CPS) and on a general purpose computer. Once trained, the model weights and parameters only consume $12.3$ MB of memory. When deploying the model to a CPS with limited resources, the model can be adjusted to satisfy the non-functional requirements of the CPS (execution time and memory utilization) with techniques such as  quantization and preprocessing pipeline tuning~\cite{b34}.

In the rest of this section, we show the results of our experiments and compare our approach to three state-of-the-art methods.

\subsection{Dataset and partitions}

We use the Carla simulator to generate images. We define city and rain as two generative factors ($F=\{f_{rain},f_{city}\}$). For gathering different values for a city generative factor, we drive a car in the Carla simulator in cities three ($SC3$), four ($SC4$), and five ($SC5$). Cities three, four, and five consist of images of rural roads, highways, and urban roads, respectively. We create four levels of rain intensity as no rain ($NR$), low rain ($LR$), moderate rain ($MR$), and heavy rain ($HR$) by sampling rain intensity from  $[0, 0]$, $[0.002 , 0.003]$, $[0.005, 0.006]$ and  $[0.008, 0.009]$, respectively. Table \ref{tab:Partition} shows the details of how each partition was gathered. 

\begin{table}[]
\centering
\begin{tabular}{|c|c|}
\hline
Partition  & Description                                                                                                                            \\ \hline
$P_1$         & \begin{tabular}[c]{@{}c@{}}Images that captured in city 3 when  \\ rain intensity is low ($SC3LR$)\end{tabular}       \\ \hline
$P_2$         & \begin{tabular}[c]{@{}c@{}}Images that captured in city 3 when  \\ rain intensity is moderate ($SC3MR$)\end{tabular}    \\ \hline
$P_3$        & \begin{tabular}[c]{@{}c@{}} Images that captured in city 4 when  \\ rain intensity is low  ($SC4LR$)                                                \end{tabular}   \\ \hline
$P_4$       & \begin{tabular}[c]{@{}c@{}} Images that captured in city 4 when  \\ rain intensity is moderate ($SC4MR$)                                                 \end{tabular}   \\ \hline 
$P_5$       & \begin{tabular}[c]{@{}c@{}} Images that captured in city 3 when  \\ rain intensity is heavy  ($SC3HR$)                                               \end{tabular}   \\ \hline
$P_6$       & \begin{tabular}[c]{@{}c@{}} Images that captured in city 3 without  \\ rain (SC3NR)                                                \end{tabular}   \\ \hline
$P_7$       & \begin{tabular}[c]{@{}c@{}} Images that captured in city 4 when  \\ rain intensity is heavy  ($SC4HR$)                                               \end{tabular}   \\ \hline
$P_8$        & \begin{tabular}[c]{@{}c@{}} Images that captured in city 4 without  \\ rain ($SC4NR$)                                                \end{tabular}   \\ \hline

$P_9$        & \begin{tabular}[c]{@{}c@{}}Images that captured in city 5 when  \\ rain intesity is low ($SC5LR$)\end{tabular}       \\ \hline
$P_{10}$       & \begin{tabular}[c]{@{}c@{}}Images that captured in city 5 when  \\ rain intensity is moderate (SC5MR)\end{tabular}    \\ \hline

$P_{11}$         & \begin{tabular}[c]{@{}c@{}}Images that captured in city 5 when  \\ rain intensity is heavy ($SC5HR$)\end{tabular}       \\ \hline
$P_{12}$        & \begin{tabular}[c]{@{}c@{}}Images that captured in city 5 without  \\ rain ($SC5NR$) \end{tabular}    \\ \hline

\end{tabular}
\caption{Data partitions that are used for creating train, calibration, and test data sets}
\label{tab:Partition}
\end{table}

Given the set of partitions ${P_1, ... ,P_{12}}$. We draw $750$ and $150$ samples from each partition $P_1,..., P_4$ to form training and calibration datasets, respectively. In total, $3000$ training data and $600$ calibration data are gathered. For generating test datasets we use all the partitions  $P_1, ... , P_{12}$. We create two datasets for evaluating the rain and city OOD reasoners. We draw $324$ samples from  partitions $P_1, ... ,P_4$ and $162$ from the rest of the partitions for evaluation of the rain reasoner (in total $2592$ samples). For analysing the city reasoner, we draw $162$ samples from partitions  $P_9, .. ,P_{12}$ and $81$ samples from other partitions (in total $1296$).  We ensured that there was an equal number of ID and OOD samples in the test sets to avoid biasing our AUROC metric for OOD reasoning~\cite{b32}.  It should be noted that training, calibration, and test datasets are mutually exclusive.

We define a set containing a pair of partitions that only differ on one generative factor  for each of the generative factors as $V_{f_{rain}}=\{(P_1,P_2),(P_3,P_4)\}$ and $V_{f_{city}}=\{(P_1,P_3),(P_2,P_4)\}$ .  We also fix dimensions $3$ and $6$ of latent space to encode rain and city generative factors ($L_{f_{rain}}=\{3\}$, $L_{f_{city}}=\{6\}$).

\subsection{Training OOD reasoners}

We train a WDLVAE as an LTN network. We re-implement code for the LTN~\cite{b30} in  PyTorch and modify it for our framework. The encoder network has four convolutional layers with depths $128/64/32/16$ with kernel size $3 \times 3 $. Each convolutional layer is followed by a batch normalization and max pooling layer with kernel size $2$.  Three fully connected layers follow the convolutional layers with sizes $2048$, $1000$, and $250$. We fix the size of latent space  to 30 ($n=30$). For the decoder, we used the mirror architecture of the encoder. 

We define four types of rules for training the WDLVAE using the functions and predicates defined in table~\ref{tab:FuncsPreds}. 
Table~\ref{tab:Trr}, shows four types of rules over partitions $P_1, .. ,P_4$  to define the loss rule for the WDLVAE.  \emph{Recloss} and  \emph{Regloss}  are necessary for training the VAE. $Adaptloss_{rain}$ ensures that the change in rain intensity level is reflected in the learned distribution in dimension $3$ of latent space. Similarly,  $Adaptloss_{city}$ ensures that the change in type of city leads to a change in the learned distribution in dimension $6$ of latent space.
$Isoloss_{rain}$ ensures that the change in rain intensity does not affect the learned distribution in other dimensions ($\overline{3}=\mathcal{L} \setminus \{3\}$) of latent space. Similarly, $Isoloss_{city}$ ensures that the change in type of city  does not affect the learned distribution in other dimensions ($\overline{6}=\mathcal{L} \setminus \{6\}$) of latent space.  The total loss is defined in equation\ref{eq:exloss} as the average over all obtained losses.
\begin{table}[]
\centering
\begin{tabular}{|c|l|}
\hline
\textbf{Rule}       & \multicolumn{1}{c|}{\textbf{Definition}}                                                \\ \hline
$Recloss$             & \begin{tabular}[c]{@{}l@{}}$\exists k \in \{1,2,3,4\}(\forall{x_k \in P_{k}}$ \\ $(Recloss(x_k,Dec_{\theta}(Samp(Enc_{\phi}(x_k))))))$
\end{tabular}                                                                                                          \\ \hline
$Regloss$             & $\exists k \in \{1,2,3,4\} \; (\forall x \in P_{k} $\\ &  $(KLU(Mean(Enc_{\phi}(x)),  Logvar(Enc_{\phi}(x)))))$ \\ \hline     

$Adaptloss_{rain}$ & \begin{tabular}[c]{@{}l@{}}$\bigwedge_{(P_k, P_k') \in \{(P_1,P_2),(P_3,P_4) \}} \sim (\forall (x, y) \in (P_k, P_{k'})$ \\    $(KLT(Mean_{3}(Enc_{\phi}(x)),Logvar_{3}(Enc_{\phi}(x)),$\\  $Mean_{3}(Enc_{\phi}(y)),Logvar_{3}(Enc_{\phi}(y))))$
\end{tabular}                                     \\ \hline

$Adaptloss_{city}$ & \begin{tabular}[c]{@{}l@{}}$\bigwedge_{(P_k, P_k') \in \{(P_1,P_3),(P_2,P_4) \}} (\sim (\forall (x, y) \in (P_k, P_{k'})$ \\    $(KLT(Mean_{6}(Enc_{\phi}(x)),Logvar_{6}(Enc_{\phi}(x)),$\\  $Mean_{6}(Enc_{\phi}(y)),Logvar_{6}(Enc_{\phi}(y)))))$
\end{tabular}                                     \\ \hline

$Isoloss_{rain} $  & \begin{tabular}[c]{@{}l@{}}$\bigwedge_{(P_k, P_k') \in \{(P_1,P_2),(P_3,P_4) \}}  (\forall (x,y) \in (P_k, P_{k'})$ \\ $(KLT(Mean_{\overline{3}}(Enc_{\phi}(x)),Logvar_{\overline{3}}(Enc_{\phi}(x)),$\\ $Mean_{\overline{3}}(Enc_{\phi}(y)),Logvar_{\overline{3}}(Enc_{\phi}(y))))) $\end{tabular} \\ \hline

$Isoloss_{city} $  & \begin{tabular}[c]{@{}l@{}}$\bigwedge_{(P_k, P_k') \in \{(P_1,P_3),(P_2,P_4) \}}  (\forall (x,y) \in (P_k, P_{k'})$ \\ $(KLT(Mean_{\overline{6}}(Enc_{\phi}(x)),Logvar_{\overline{6}}(Enc_{\phi}(x)),$\\ $Mean_{\overline{6}}(Enc_{\phi}(y)),Logvar_{\overline{6}}(Enc_{\phi}(y))))) $\end{tabular} \\ \hline

\end{tabular}
\caption{Rules that are used for training the WDLVAE}
\label{tab:Trr}
\end{table}

\begin{equation}
    \begin{gathered}
        Loss(P_1,P_2,P_3,P_4)= \\
        \frac{1}{6}[Recloss+Regloss+
        \\\sum_{f_i \in \{f_{rain},f_{city}\}} (Adaptloss_{f_i}+Isoloss_{f_i})]
    \end{gathered}
    \label{eq:exloss}
\end{equation}

\subsection{Run-time OOD reasoning}

We trained the WDLVAE on calibration data and then applied the k-means algorithm on the obtained distributions for dimensions $3$ and $6$ for the generative factors rain and city respectively. Two clusters are formed in these two dimensions as the calibration data contains two observed values for each generative factor. Next, we approximate two Gaussian mixture models in dimensions $3$ and $6$ by using the obtained centers of clusters in each of the aforementioned dimensions. Based on whether or not the test sample is derived from an obtained mixture model in the selected dimensions, we identify the sample as OOD with respect to the generative factors rain and city.

\subsection{Competing methods}

We evaluate our approach by comparing it alongside three state-of-the-art approaches~\cite{b6} \emph{HPVAE},~\cite{b7} \emph{PVAE} and~\cite{b8} \emph{AVVAE}. 
We keep the VAE architectures the same as the original study for each of the three methods. We also tune hyper parameters $\beta $ and $n$ as advised in these papers. We fixed dimension $0$ for both $PVAE_{rain}$ and $PVAE_{city}$ as advised in the implementation of this approach. Since the dimensions responsible for encoding rain and city are unknown in the \emph{AVVAE} approach, we use the same approach as  \emph{HPVAE} to identify them as explained in section \ref{sc:Related}. Also, we train separate VAEs for each generative factor (rain, and city) for the \emph{AVVAE} and \emph{PVAE} methods. The reason is that the performance of these approaches decreases significantly for more than one factor. Table \ref{tab:Hyper} shows the selected hyperparameters and the approach for tuning them.

\begin{table}[]
\centering
\begin{tabular}{|c|c|c|c|c|}
\hline
\textbf{Approach} & \textbf{$\beta$} & \textbf{\begin{tabular}[c]{@{}c@{}}Latent\\ space size\\ ($\mathcal{L}$)\end{tabular}} & \textbf{\begin{tabular}[c]{@{}c@{}}Selected \\ latent \\ dimension\end{tabular}} & \textbf{\begin{tabular}[c]{@{}c@{}}Hyper parameter \\ tuning approach\end{tabular}} \\ \hline

$HPVAE$         & 3.13             & 152                                                                                    & 125                                                                           &  \begin{tabular}[c]{@{}c@{}}BO over \\ $\mathcal{L}=[30,200]$\\ and $\beta=[2,10]$\end{tabular}                                           \\ \hline
$AVVAE_{rain}$   & 6                & 100                                                                                    & 15                                                                            & GS over  \\ & & & &
$\beta=[1 ,2 ,4 ,6 ,8 ,16]$

\\ \hline
$AVVAE_{city}$   & 4                & 100                                                                                    & 50                                                                            & GS over  \\ & & & &
$\beta=[1 ,2 ,4 ,6 ,8 ,16]$                                                                        \\ \hline
$PVAE{rain}$      & 4                & 128                                                                                    & 0                                                                             &  GS over  \\ & & & &          $\beta=[1,10]$                                                             \\ \hline
$PVAE_{city}$     & 10               & 128                                                                                    & 0                                                                             & GS over \\ & & & & $\beta=[1,10]$     \\ \hline
\end{tabular}
\caption{Selected hyper parameters for competing approaches. BO is Bayesian optimization and GS is grid search}
\label{tab:Hyper}
\end{table}

\subsection{The effect of disentanglement constraints}


To demonstrate the effect of the introduced disentanglement constraints (\emph{Adaptloss}, \emph{Isoloss}) on the learned latent space, we provide qualitative analysis (latent visualization) and quantitative analysis (mutual information).


\textbf{Latent visualization:} Figure \ref{fig:rainreason} shows the selected dimension for the rain  reasoner obtained by four methods. In figures \ref{fig:WDrlat3}-\ref{fig:wrainl15} different levels of rain intensity are shown by different colors. Dark green, green, dark red, and red show low rain ($LR$), moderate rain ($MR$), no rain ($NR$), and heavy rain ($HR$), respectively. The shape of points shows different values for the city factor. The square, circle, and triangle markers represent city three ($SC3$), city four ($SC4$), and city five ($SC5$), respectively. As shown in figure \ref{fig:WDrlat3} latent dimension $3$ of the \emph{WDLVAE} successfully separates in-distribution samples with values $LR$ and $MR$ for rain intensity from OOD samples $NR$ and $HR$ by using $Adaptloss_{rain}$ during training. The $Adaptloss_{rain}$ rule leads to learning a mixture of Gaussian distribution that can more effectively estimate the distribution of the rain generative factor. As shown in figure \ref{fig:JorVAErain} although the \emph{HPVAE} model separates $NR$ OOD samples from in-distribution samples, the latent dimension $125$ is not sensitive enough to changes in the rain generative factor. Figures \ref{fig:prainl0} of latent dimension $0$ of the $PVAE_{rain}$ and  \ref{fig:wrainl15} of latent dimension $15$ of the $AVVAE_{rain}$ show that these models are unable to identify different values of the rain generative factor. The poor performance of the three comparison methods is mainly because these approaches do not consider the effect of change in the city generative factor in the distribution learned for the rain factor. Using $Isoloss_{city}$ helps the  \emph{WDLVAE} to alleviate the effect of change in the city generative factor on latent dimension $3$ of latent space.

Figure \ref{fig:cityreasoner} depicts the selected dimension for the city reasoner generated by the four methods. In  figures \ref{fig:WDclat6}-\ref{fig:wcityl50}, different types of city are shown in different colors. Dark green and green show in-distribution samples captured in cities three ($SC3$) and four ($SC4$), respectively. The red color shows the OOD city five ($SC5$). The shape of the points shows different values for rain intensity levels. The square, circle, down-pointing triangle, and right-pointing triangles represent $LR$, $MR$, $HR$, and $NR$, respectively. As shown in figure \ref{fig:WDclat6}, latent dimension $6$  successfully separates in-distribution samples  $SC3$ and $SC4$ from OOD samples $SC5$. In the latent dimension $6$, a Gaussian mixture distribution is learned using $Adaptloss_{city}$ during training. This distribution is more suitable for learning city latent representations than uni-modal or multi-modal Gaussian distributions. As shown in figure \ref{fig:JorVAEcity}, latent dimension $125$ also captures the distribution of the city factor alongside the rain factor. The \emph{HPVAE}  model did not use fixed dimensions for each generative factor. In addition, it was not able to isolate the effect of the change in the rain factor on the city distribution during training. As a result, the model learned the same dimension to encode the distribution of both rain and city factors. The poor representation of the city generative factor by the \emph{HPVAE} shows the importance of using $Isoloss_{rain}$  during training. For \ref{fig:pcityl0} and \ref{fig:wcityl50}, although for the city  generative factor  different values of city  are separated to some degree, dimensions $0$ and $50$ are not sensitive enough to change in the city  factor. In other words, a uni-modal Gaussian distribution can not properly capture the distribution of different values of city factor. The aforementioned problem is prevented in the \emph{WDLVAE} by using the $Adaptloss_{city}$ rule during training.

\textbf{Mutual information}: To show the effect of $Adaptloss$ and $Isoloss$ on the structure of latent dimensions assigned to generative factors, we identify the most informative and second most informative latent dimensions based on mutual information~\cite{b28} between each generative factor and latent dimensions in latent space. Mutual information shows the amount of information that dimension $L_l \in \mathcal{L}$ in latent space has about generative factor $f_i$. Mutual information is calculated as $I(f_i ; L_l) = H(f_i) – H(f_i | L_l)$ where $I$ is mutual information. $H(f_i)$ is the entropy of a generative factor and $H(f_i | L_l)$ is the conditional entropy of a generative factor $f_i$ conditioned on latent dimension $L_l$~\cite{b28}. 

As shown in table \ref{tab:MIG} for the \emph{WDLVAE} the most informative latent dimension for the rain factor is the same as the fixed latent dimension. Using $Adaptloss$ leads to sensitivity of latent dimensions $3$ and $6$ to rain and city factors, respectively. Also, there is a considerable gap between mutual information between the first and second most informative latent dimensions for each factor. Using $Isoloss$ leads to encoding information about rain and city factors mostly in dimensions $3$ and $6$ of latent space, respectively. In addition, since mutual information is calculated using the training set and selection of latent dimension is done by calibration set in the \emph{HPVAE} and the \emph{AVVAE}, there is a mismatch between selected dimensions. In other words, change in data can lead to a different encoding of factors in the latent dimension. Therefore correspondence between reasoner dimensions and  factors must be established during training explicitly.

As shown in the table, None of the competing approaches shows the \emph{WDLVAE}'s level of adaptability and isolation for the selected latent dimensions.

\begin{table}[]
\centering
\begin{tabular}{|c|c|c|c|c|c|c|}
\hline
\textbf{Factor}       & \textbf{Approach} & \textbf{$L_m$} & \textbf{$M_m$} & \textbf{$L_{sm}$} & \textbf{$M_{sm}$} & \textbf{$E_L$} \\ \hline
\multirow{4}{*}{rain} & \emph{WDLVAE}          & 3              & 0.36                                     & 21                & 0.01              & 3              \\ \cline{2-7} 
                      & $HPVAE$           & 125            & 0.28                                     & 49                & 0.28              & 125            \\ \cline{2-7} 
                      & $AVVAE_{rain}$    & 28             & 0.01                                     & 30                & 0.01              & 15             \\ \cline{2-7} 
                      & $PVAE_{rain}$     & 0              & 0.02                                     & 116               & 0.01              & 0              \\ \hline
\multirow{4}{*}{city} & $WDLVAE$          & 6              & 0.69                                     & 3                 & 0.02              & 6              \\ \cline{2-7} 
                      & $HPVAE$           & 49             & 0.16                                     & 125               & 0.15              & 125            \\ \cline{2-7} 
                      & $AVVAE_{city}$    & 94             & 0.01                                     & 93                & 0.01              & 50             \\ \cline{2-7} 
                      & $PVAE_{city}$     & 0              & 0.69                                     & 15                & 0.38              & 0              \\ \hline
\end{tabular}
\caption{Expected dimension $E_L$, most informative ($L_m$) and second most informative ($L_{sm}$) latent variable and their mutual information gain $M_m$ and $M_{sm}$}
\label{tab:MIG}
\end{table}

\begin{figure*}%
\centering
\begin{subfigure}{.5\columnwidth}
\includegraphics[width=\columnwidth]{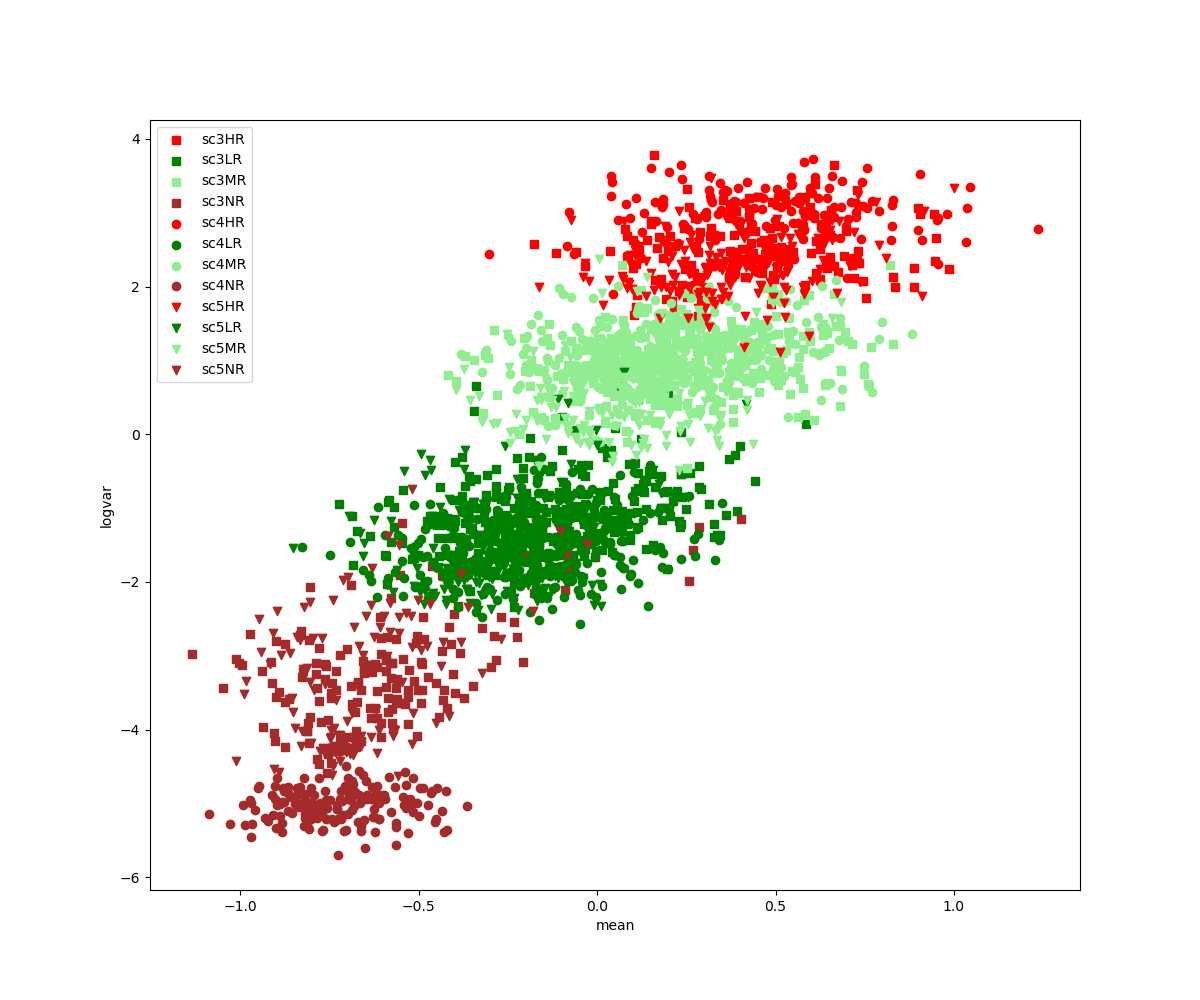}%
\caption{}%
\label{fig:WDrlat3}%
\end{subfigure}\hfill%
\begin{subfigure}{.5\columnwidth}
\includegraphics[width=\columnwidth]{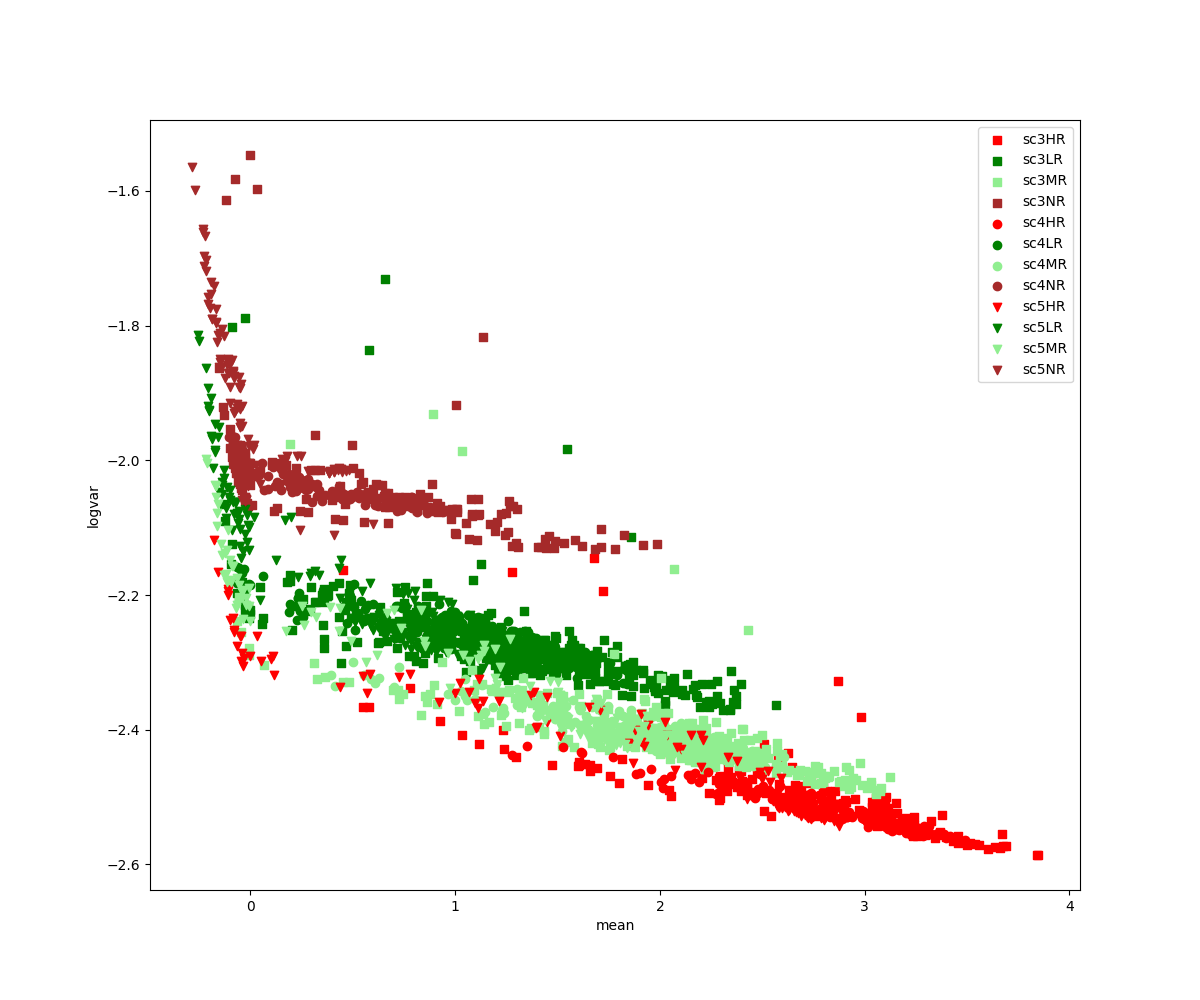}%
\caption{}%
\label{fig:JorVAErain}%
\end{subfigure}\hfill%
\begin{subfigure}{.5\columnwidth}
\includegraphics[width=\columnwidth]{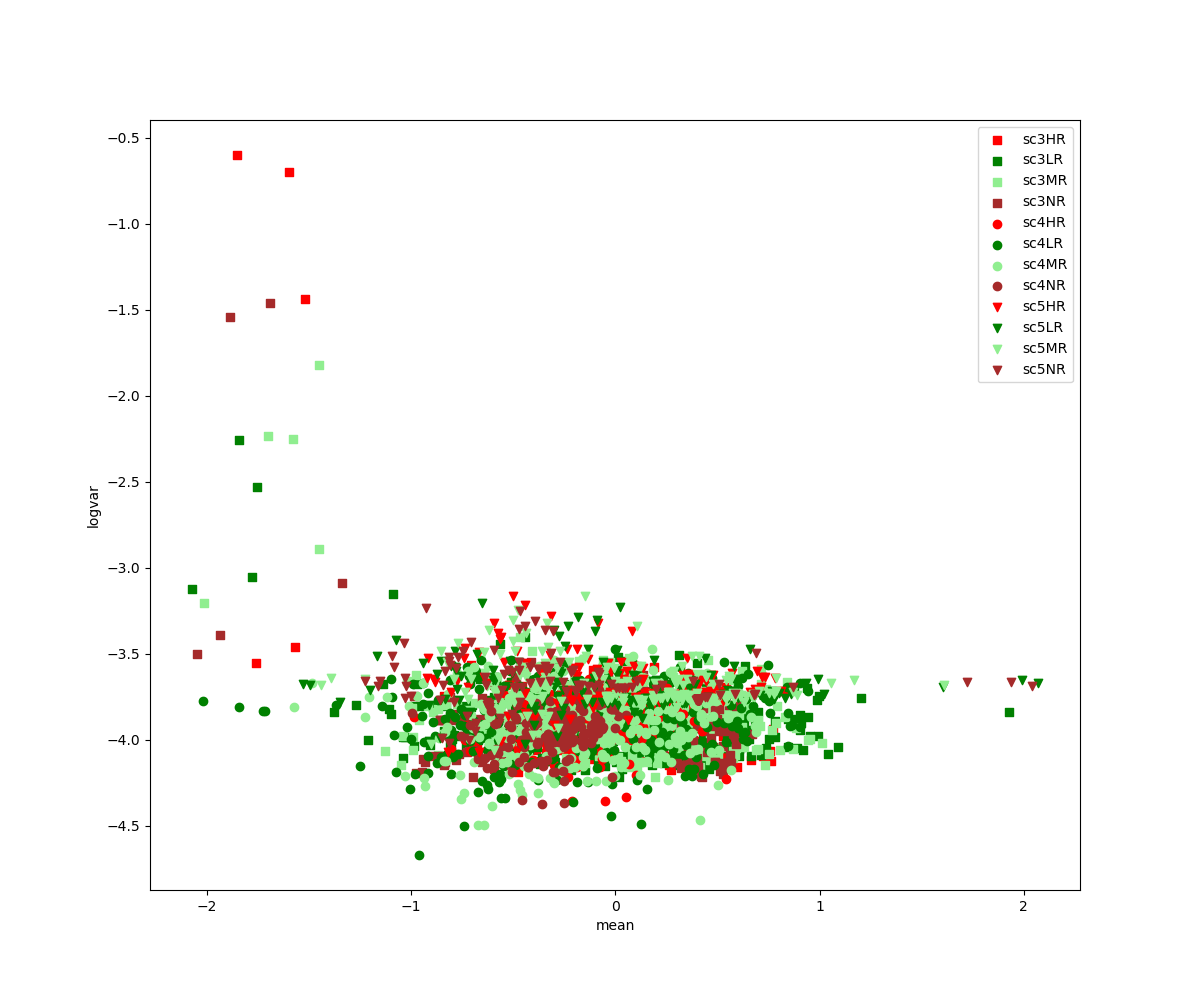}%
\caption{}%
\label{fig:prainl0}%
\end{subfigure}\hfill%
\begin{subfigure}{.5\columnwidth}
\includegraphics[width=\columnwidth]{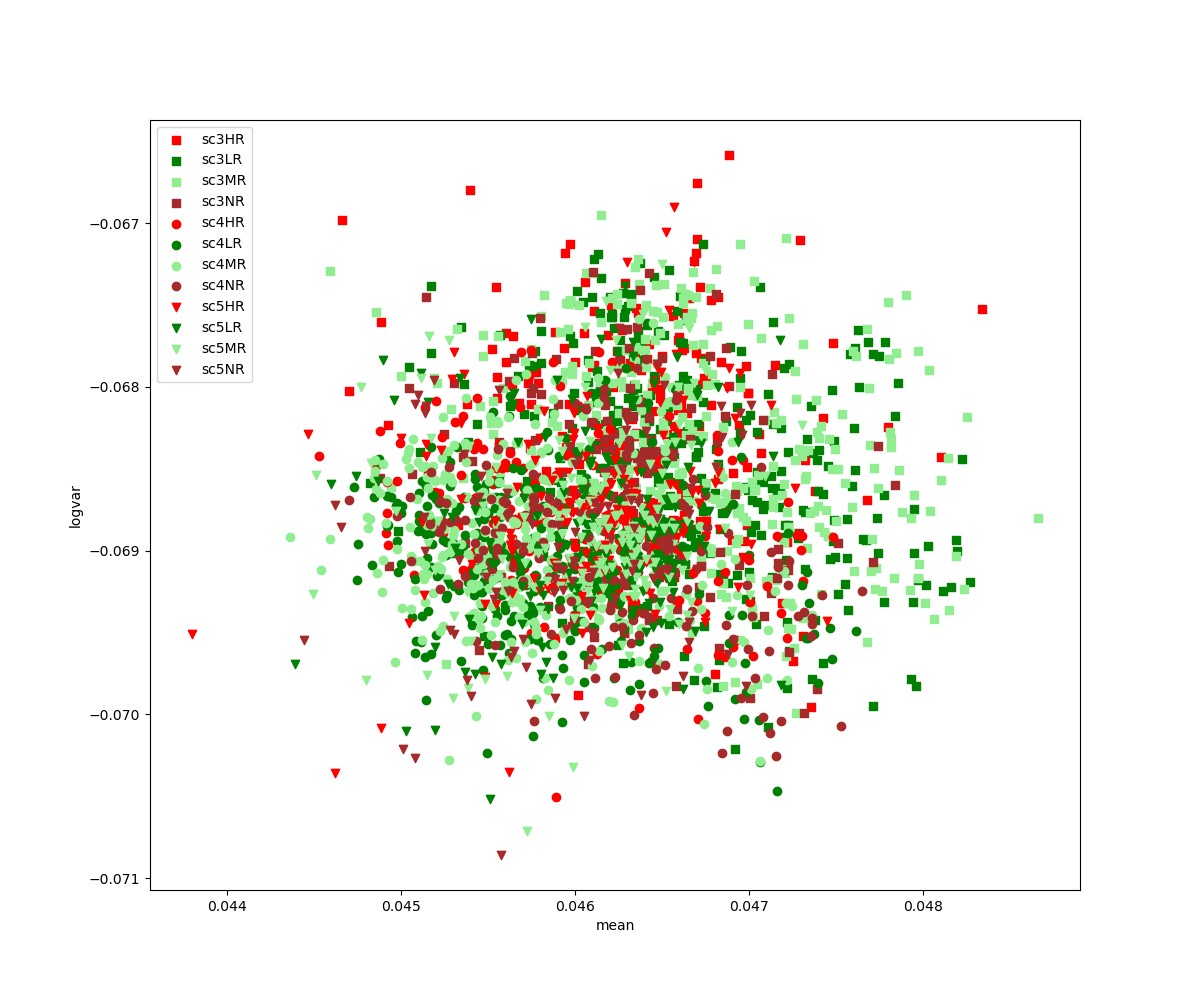}%
\caption{}%
\label{fig:wrainl15}%
\end{subfigure}\hfill%
\caption{Latent dimensions that are used as rain reasoner obtained by (\ref{fig:WDrlat3}) WDLVAE, (\ref{fig:JorVAErain}) HPVAE, (\ref{fig:prainl0}) PVAE, (\ref{fig:wrainl15}) AVVAE approaches}%
\label{fig:rainreason}%
\end{figure*}

\begin{figure*}%
\centering
\begin{subfigure}{.5\columnwidth}
\includegraphics[width=\columnwidth]{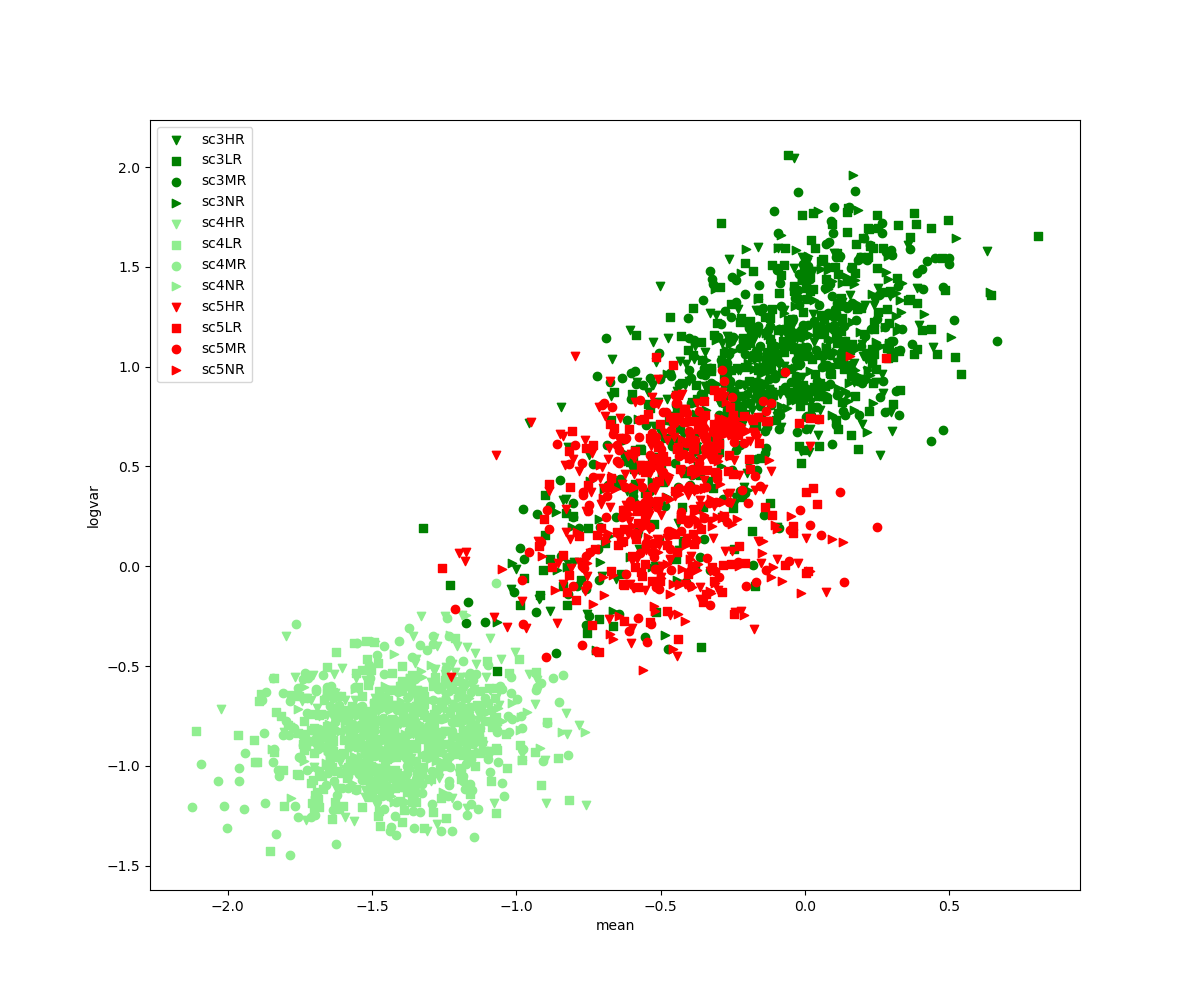}%
\caption{}%
\label{fig:WDclat6}%
\end{subfigure}\hfill%
\begin{subfigure}{.5\columnwidth}
\includegraphics[width=\columnwidth]{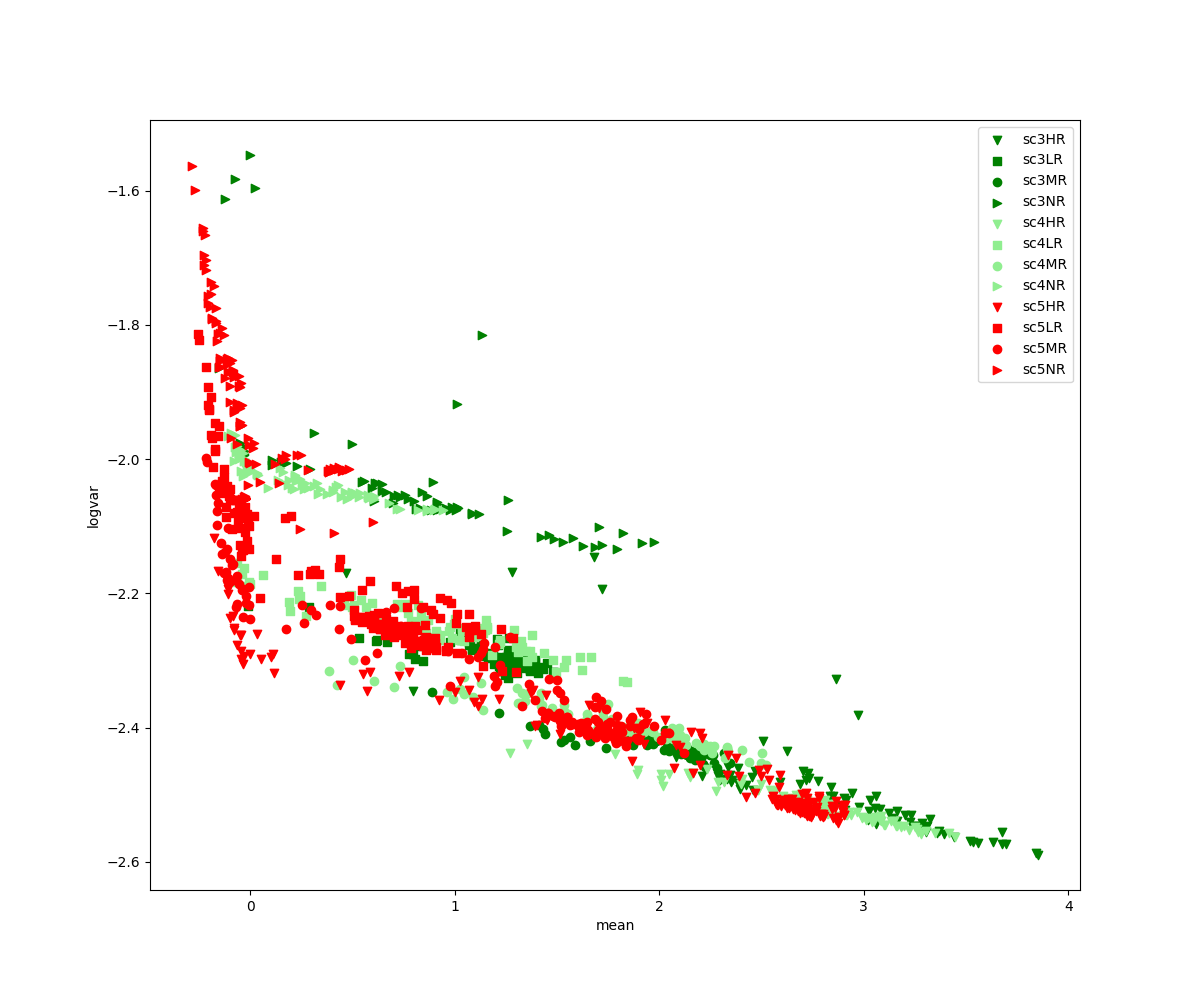}%
\caption{}%
\label{fig:JorVAEcity}%
\end{subfigure}\hfill%
\begin{subfigure}{.5\columnwidth}
\includegraphics[width=\columnwidth]{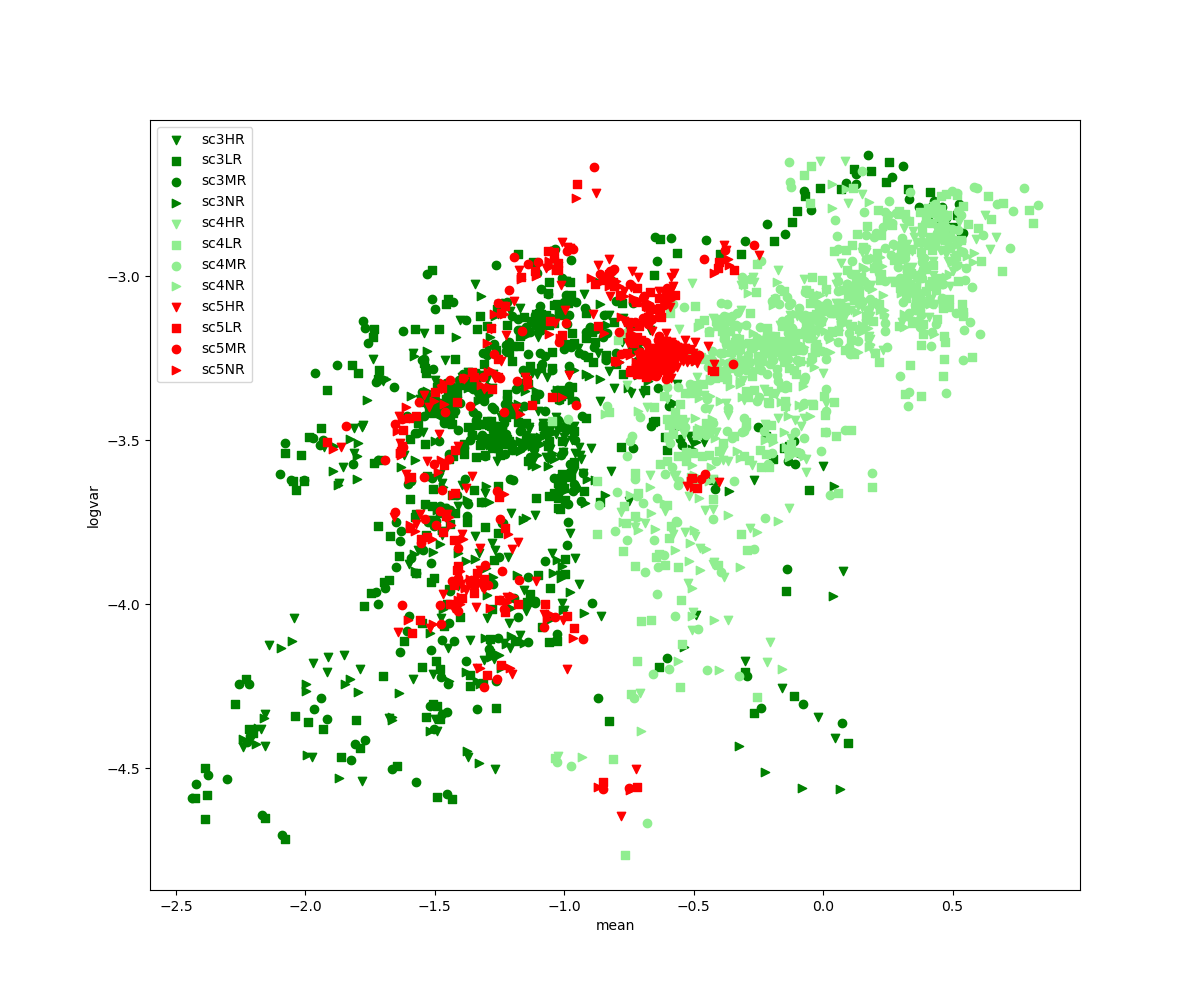}%
\caption{}%
\label{fig:pcityl0}%
\end{subfigure}\hfill%
\begin{subfigure}{.5\columnwidth}
\includegraphics[width=\columnwidth]{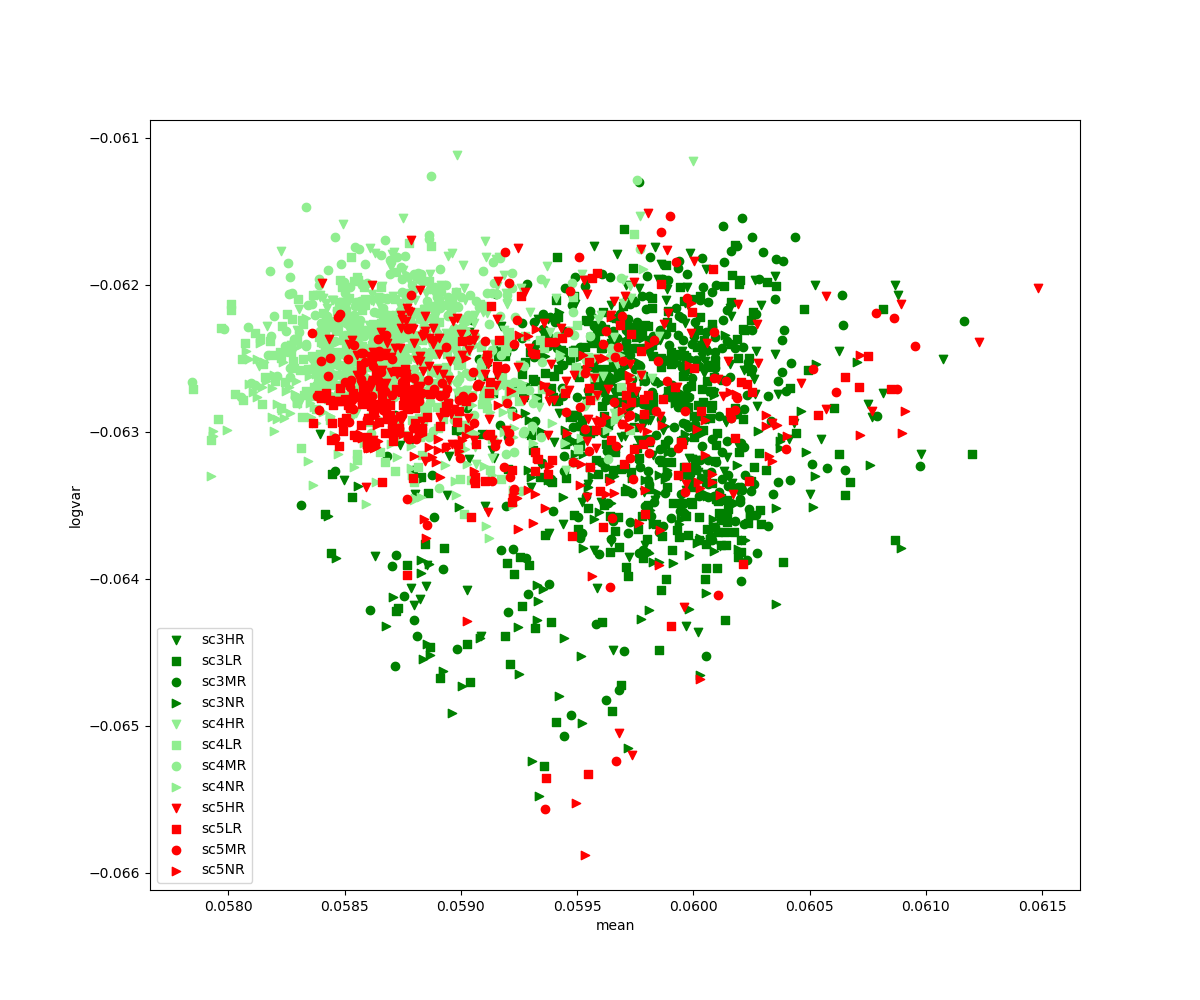}%
\caption{}%
\label{fig:wcityl50}%
\end{subfigure}\hfill%
\caption{Latent dimensions that are used as city  reasoner obtained by (\ref{fig:WDclat6}) WDLVAE, (\ref{fig:JorVAEcity}) HPVAE, (\ref{fig:pcityl0}) PVAE, (\ref{fig:wcityl50}) AVVAE approaches }%
\label{fig:cityreasoner}%
\end{figure*}

\begin{figure*}%
\centering
\begin{multicols}{2}
   \includegraphics[width=0.8\linewidth]{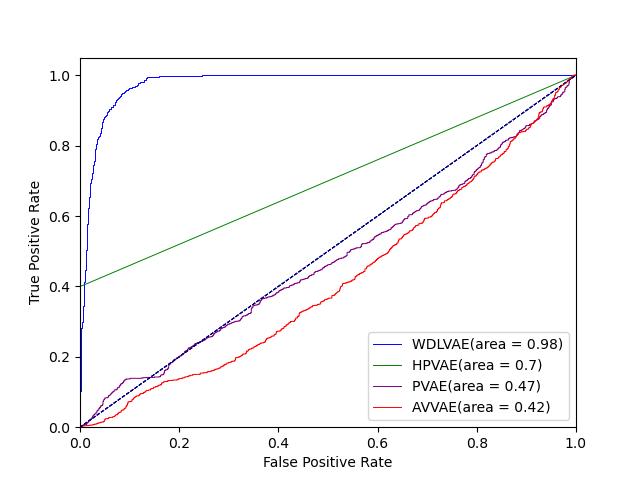}
   \caption{AUROC for rain\\
   reasoner}
    \label{fig:FinAUROCrain}
    \includegraphics[width=0.8\linewidth]{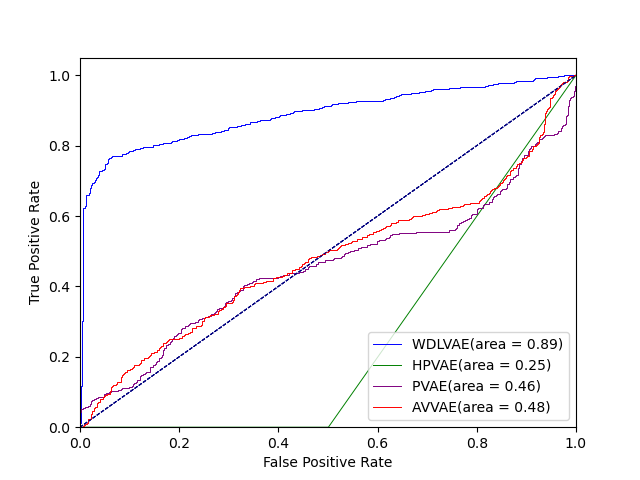}
    \caption{AUROC for city\\
   reasoner }
   \label{fig:FinAUROCcity}
    \end{multicols}

\end{figure*}

\subsection{OOD reasoning performance:}
We have demonstrated the \emph{WDLVAE}'s ability to obtain a  disentangled latent space, however, our end goal is to use the selected dimensions to perform OOD reasoning.  Thus, we compare its OOD reasoning performance with that of the \emph{PairVAE} and the \emph{AVVAE}.  These two approaches have not yet been applied to OOD reasoning, so we take their selected latent dimensions for the rain and city factors and perform the same run-time OOD reasoning steps as the \emph{WDLVAE} for a fair comparison. 


 We use AUROC curves to show the performance of rain and city reasoners. AUROC curves depict the tradeoff between true positive rate ($TPR=\frac{TP}{TP+FN}$) and false positive rate ($FPR=\frac{FP}{TN+FP}$). Where $TP$, $FP$,$TN$, and $FN$ are true positives, false positives, true negatives, and false negatives, respectively. 
Figures \ref{fig:FinAUROCrain} and \ref{fig:FinAUROCcity} show the AUROC curves for rain and city factors.
Our framework has an AUROC of $0.98$ and $0.89$ for rain and city reasoners, respectively. The \emph{AVVAE} and \emph{PVAE} methods perform poorly (AUROC $\sim$ 0.5) for both generative factors. This poor performance is mainly because these methods cannot handle changes in other factors at the same time as a factor of interest and cannot represent complex data distributions properly, especially for the city generative factor. The \emph{HPVAE} performs better than the other two methods for the rain factor (\emph{AUROC}$=0.7$). However, in this method, the selection of dimensions is postponed to post-training causing the selected dimension $125$ to differ from the real dimension that encodes the city type factor for the test dataset. Therefore, poor performance is shown for the city factor (\emph{AUROC}$=0.25$).

\section{Conclusion}

In this paper, we propose a framework for OOD reasoning for complex datasets with generative factors defined abstractly and have continuous domains. We use LTN to distill disentanglement knowledge during training in VAE weights. We introduce $Adaptloss$ constraints to ensure the generative factor change will reflect on the learned distributions of the corresponding dimensions. We also introduce $Isoloss$ to ensure a change in a given generative factor does not affect other dimensions of latent space. We then used selected disentangled dimensions as OOD reasoners to identify OOD behavior with respect to a given factor.
We evaluate our approach on the Carla dataset. We show the importance of the disentanglement constraints $Adaptloss$  and $Isoloss$ in achieving disentanglement for two factors of rain and city by visualization and mutual information. We also evaluate the performance of the OOD reasoner by AUROC curves for each reasoner. In the future, we want to generalize this approach to a time-stamped, real-world dataset such as nuScenes~\cite{b31} using a temporal logic extension of LTN.

\end{document}